\def\year{2019}\relax
\documentclass[letterpaper]{article} 
\usepackage{aaai19}  
\usepackage{times}  
\usepackage{helvet} 
\usepackage{courier}  
\usepackage[hyphens]{url}  
\usepackage{graphicx} 
\usepackage{graphicx}  
\newcommand{\citet}[1]{\citeauthor{#1} \shortcite{#1}}
\newcommand{\citep}{\cite}

\usepackage{amssymb,amsmath}
\usepackage{balance}
\usepackage{booktabs}
\usepackage[font=footnotesize,labelformat=simple]{subcaption}
\usepackage[font=footnotesize]{caption}

\newcommand{\figref}[1]{Fig.~\ref{#1}}

\newcommand{\xxnote}[3]{}
\ifx\hidenotes\undefined
  \usepackage{color}
  \renewcommand{\xxnote}[3]{\color{#2}{#1: #3}}
\fi

\title{Towards Effective Human-AI Teams: The Case of Collaborative Packing}
\author{Gilwoo Lee, Christoforos Mavrogiannis, Siddhartha S. Srinivasa\\ 
Paul G. Allen School of Computer Science \& Engineering\\
University of Washington\\
3800 E Stevens Way NE\\
Seattle, WA 98195-2350, USA\\
$\lbrace$gilwoo, cmavro, siddh$\rbrace$ @cs.uw.edu
}
 
\begin{document}

\maketitle

\begin{abstract}
We focus on the problem of designing an artificial agent (AI), capable of assisting a human user to complete a task. Our goal is to guide human users towards optimal task performance while keeping their cognitive load as low as possible. Our insight is that doing so requires an understanding of human decision making for the task domain at hand. In this work, we consider the domain of collaborative packing, in which an AI agent provides placement recommendations to a human user. As a first step, we explore the mechanisms underlying human packing strategies. We conducted a user study in which 100 human participants completed a series of packing tasks in a virtual environment. We analyzed their packing strategies and discovered spatial and temporal patterns, such as that humans tend to place larger items at corners first. We expect that imbuing an artificial agent with an understanding of this spatiotemporal structure will enable improved assistance, which will be reflected in the task performance and the human perception of the AI. Ongoing work involves the development of a framework that incorporates the extracted insights to predict and manipulate human decision making towards an efficient trajectory of low cognitive load and high efficiency. A follow-up study will evaluate our framework against a set of baselines featuring alternative strategies of assistance. Our eventual goal is the deployment and evaluation of our framework on an autonomous robotic manipulator, actively assisting users on a packing task.
\end{abstract}

\section{Introduction}

We consider the general scenario in which a human and an artificial agent (AI) are collaborating to jointly complete a task. Depending on the domain, it is often true that the capabilities of the human and the robot may greatly differ. The AI agent may often possess superior computation capabilities, whereas the human agent may have superior perceptual and control abilities. An emerging area of research looks at the development of frameworks that would enable effective combined performance by leveraging the strengths of both parties, while ensuring human comfort. Our insight is that to achieve this goal, the AI needs to reason about the decision making of its human counterpart. At times, it may need to intervene to guide their behavior towards efficiency. In many cases, this intervention may not be realized physically due to hardware or software limitations. In such cases, it is critical that the AI conveys its intentions implicitly through its actions.

\begin{figure}
\centering
\includegraphics[width=\linewidth]{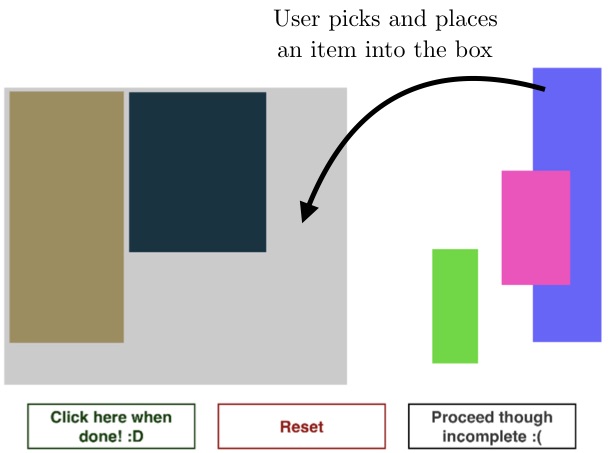}
\caption{We study human packing strategies. In our study interface, the user sequentially drags and drops a set of objects (shown on the right side) into a container (depicted on the left side). When done, the user may proceed to the next task by clicking the "done" button (bottom left). The user is also given the options of resetting the task to its initial state ("Reset" button), and proceeding without completing the task (right button). \label{fig:interface}}
\end{figure}

In this work, we consider a scenario of collaborative packing, in which a human places a set of objects in a container under the assistance of a recommender system. This application is of particular relevance these days, given the increasing presence of AI systems in logistics \citep{McKinsey}. Achieving adequate spatial efficiency in packing is an anecdotally hard problem for non-expert humans, whereas manipulation is still a big challenge for robots. On the other hand, AI systems feature superior long-term planning capabilities, whereas humans are equipped with unparalleled manipulation capabilities. An effectively combined collaborative effort, leveraging the strengths of both could help achieve increased overall performance.

As a first step to approach the outlined vision, we seek to understand the domain of packing, focusing on the strategies that humans employ when faced with completing a packing task. To do so, we conducted an online user study in which we asked human subjects to complete a series of packing tasks in a virtual environment. Each task involved the placement of a different set of 2-dimensional objects inside a packing container. Our findings suggest that human packing strategies in this domain can largely be classified into a set of distinct categories corresponding to different spatiotemporal patterns of placement. We discuss our findings and the ongoing development of a planning-under-uncertainty framework targeted towards ensuring improved efficiency and low cognitive load for humans in collaborative packing scenarios.

\section{Related work}

The concept of human-AI teaming is gaining popularity, as combining the strengths of humans and AI systems opens promising avenues for a variety of fields and applications \citep{Kamar12,Lasecki12b,Bayati14,Kamar16ijcai}. Naturally, the problem of enabling seamless, natural, and efficient collaboration in human-AI teams has received considerable attention over the recent years, with researchers focusing on different aspects of the interaction, such as the powerful communicative impact of actions performed in a shared context \citep{Liang19} or the tradeoffs between performance gains and compatibility with existing human mental models \citep{bansal2019updates}.

For a series of applications, transferring the benefits of human-AI teaming in the physical environment implies embodiment in robot platforms. Human-robot teaming has a unique potential for a variety of applications, given that robots can be both intelligent \emph{and} physically capable. Thus the combination of their capabilities with those of humans may result in performance standards that neither party could otherwise achieve in isolation~\cite{hoffman2004collaboration}. Notable examples include fast task completion in sequential manipulation tasks~\cite{hayes2015effective}, and improved performance through intelligent resource allocation to human participants~\cite{jung2018robot}.

A common complication in such applications is that explicit communication between the human and the robot is often not feasible, effective or desired. Therefore, in order to be of assistance, the robot needs to infer the intentions of the user implicitly, through observation of their actions, and clearly communicate its own intentions, through its own actions \citep{knepper_hri_2017}. A typical paradigm of particular relevance in this domain is shared autonomy, in which a robot assists a human user in completing their task. In a variety of applications, it has been shown that inferring and adapting to human intentions is positively perceived by users and effective \citep{Dragan13blending,kuderer14_wheelchair,Gopinath17,javdani18}. Furthermore, understanding the mechanisms underlying human decision making in a particular domain is shown to yield performance improvements and positive impressions in joint tasks \citep{Nikolaidis13,Nikolaidis15}. Finally, explicitly collaborative tasks such as collaborative manipulation \citep{DraganAuR14,nikolaidis17}, and assembly \citep{Knepper15recovering} or implicitly collaborative tasks such as social navigation \citep{Mavrogiannis19} benefit significantly by the incorporation of models of human inference.

In this work, we consider a joint task (packing), performed in collaboration between a human and an AI agent. We also consider a setting of implicit communication, in the sense that human intentions are not directly observed and need to be inferred. Our first step towards approaching this scenario is to understand the domain by collecting and analyzing human data.

\section{Study design}

Our study was conducted online on an interactive web application. Participants were recruited online, through the Amazon Mechanical Turk platform \citep{Buhrmester11mturk}. Upon providing consent, each participant was assigned the same set of 65 packing tasks, presented in a random order. These tasks involved the placement of sets of 4-8 rectangular objects of different sizes inside a rectangular container of fixed size.

\subsection{Interface}

The web interface depicts a set of rectangular objects of various dimensions, alongside a rectangular container, from a top view (see \figref{fig:interface}). Participants were instructed to sequentially place all of these objects at locations of their choice, inside the container. Once an object is placed inside the container, it cannot be moved -- thus participants are forced to judiciously decide on the placements of their objects. The interface comprises three buttons: (a) a button for proceeding to the next task (shown at the bottom left); (b) a reset button (middle), useful in cases where participants' decisions did not allow them to put all object to the container; (c) a button which allowed participants to proceed to the next task without completing the current one (right). Since at this stage we were interested in understanding the domain of packing rather than participants' performance, we gave users the option of resetting a task to its initial state by hitting the ``Reset" button. We also gave participants the option to skip a task if they decided to but we disincentivized this option by placing a sad face on the corresponding button. Similarly, we incentivized completion by placing a happy face on the ``Done" button.

\begin{figure*}
\centering
\begin{subfigure}[b]{0.30\linewidth}
\centering
\includegraphics[width=0.8\linewidth]{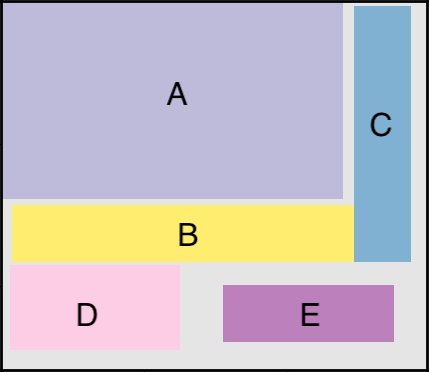}
\vspace{0.5cm}
\caption{Example packing instance\label{fig:study_box_visualization}}
\end{subfigure}
~
\begin{subfigure}[b]{0.30\linewidth}
\includegraphics[width=1\linewidth]{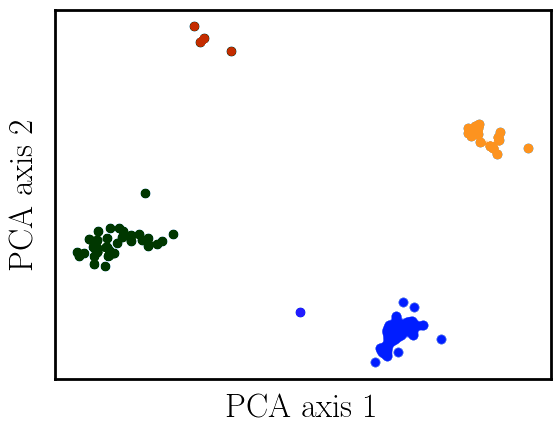}
\caption{Spatial clusters\label{fig:study_box_cluster}}
\end{subfigure}
~
\begin{subfigure}[b]{0.30\linewidth}
\includegraphics[width=1\linewidth]{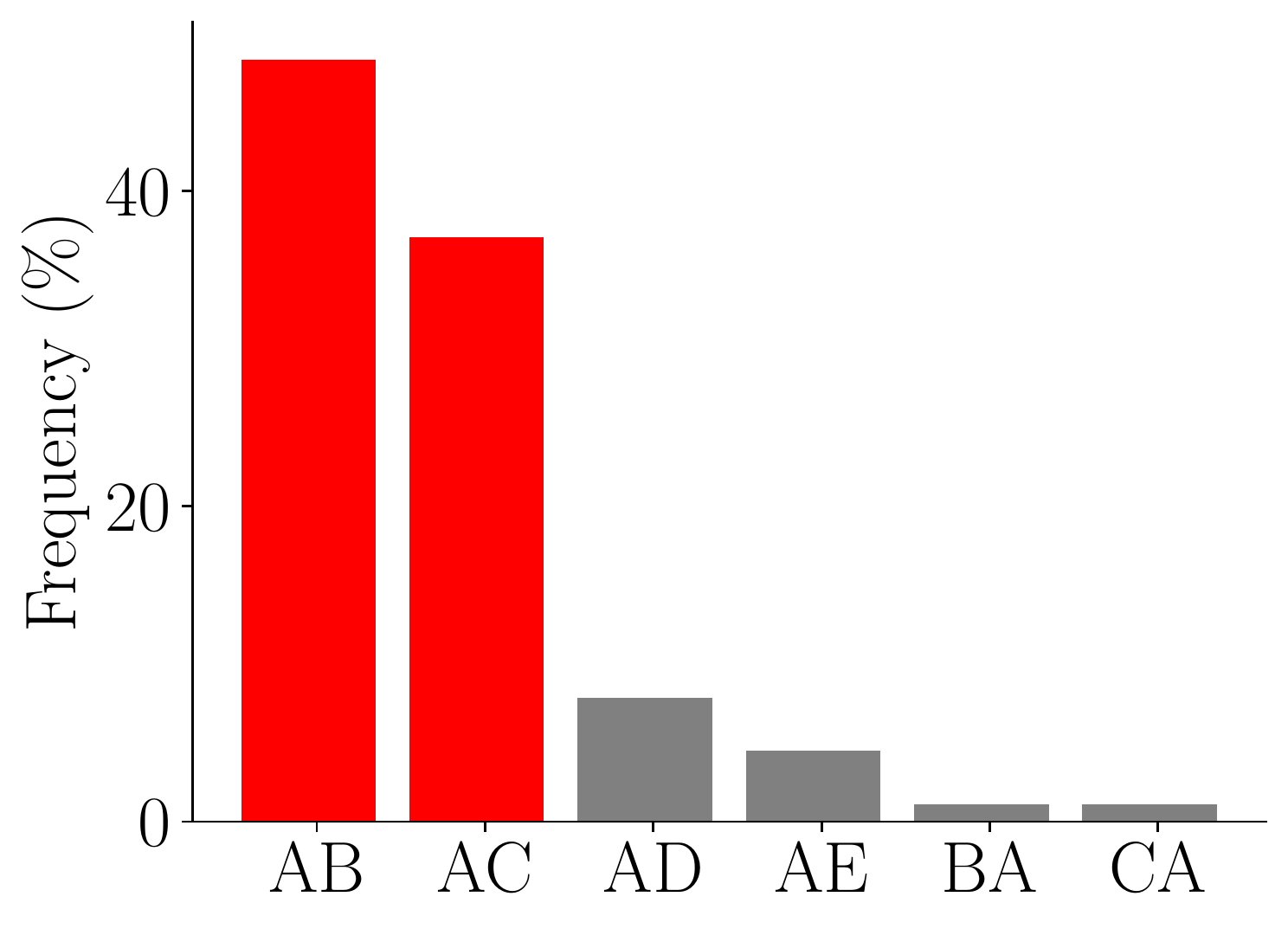}
\caption{Temporal patterns of first 2 items\label{fig:study_frequency}}
\end{subfigure}
\caption{Highlights from our study. We found a few strong spatiotemporal patterns in people's packing styles, such as placing  large items into corners or packing larger items before smaller ones. (a) An example packing instance by a user. This person packed in the order of A-C-B-D-E. (b) Spatial clusters of configurations for the items in (a). Because people tend to put larger items into corners, the final configurations can be clustered into a few spatial patterns. For this task, 4 strong spatial patterns are shown. (c) Temporal patterns of the first two items for this task. Nearly everyone chose the largest item (A) first, and 85\% of them picked one of the two second largest items (B or C).\label{fig:results}}
\end{figure*}

\subsection{Generation of Packing Tasks}

The complexity of a packing task depends on the relative sizes of objects among themselves and with respect to the container. In practice, these relationships yield different tolerance requirements in the object placements. The smaller the tolerances, the higher the amount of precision required to ensure a collision-free placement, and thus the more complex the packing task becomes.

While we can generate arbitrarily easy or complex packing tasks by designing a large container and many small items that could be packed in many different ways, such problem instances would not help us observe any discernible patterns that humans may naturally have.
In order to identify spatiotemporal patterns in packing, we have designed our packing tasks such that each task satisfies the following conditions:
\begin{itemize}
    \item At least 70\% of the container is filled with the items.
    \item There is a finite number of clusters of spatially feasible solutions.
\end{itemize}
By committing to these conditions, we constrain the tasks to be doable with a finite number of qualitatively equivalent object placements. Our expectation was that even under the constrained setting of finite spatially feasible solutions, innate human packing styles and preferences would still manifest themselves. In particular, we expected that human packing strategies would show strong inclinations towards distinct classes of spatiotemporal placements.

In order to generate packing tasks that satisfy the above conditions, we fix the size of the container, randomly generate items of various sizes, and then attempt to place them. For any resulting placement, if more than 70\% of the container is filled with 4-8 items, then we test if the second condition is met. To do so, we empty the container and attempt to place the same items in different ways. If we can generate more than 50 different collision-free configurations, then we run a Principal Components Analysis (PCA) \cite{Jolliffe} on the configurations. Each configuration is represented with a vector $x\in\mathbb{R}^{2n}$, stacking the Cartesian coordinates of all $n$ items. We take the first two dimensions of the PCA projection, visually check for discernible clusters as in \figref{fig:study_box_cluster}, and keep the task only if such clusters are found. In total, we have generated 65 tasks of varying complexity with the following distribution: 17 tasks of 4 objects, 15 tasks of 5 objects, 20 tasks of 7 objects, and 10 tasks of 8 objects.

\section{Dataset \& Analysis}

We had 100 participants (34 female, 66 male), recruited through the Amazon Mechanical Turk platform. The participants were between 18 and 65 years old ($M = 32.08$, $SD = 8.87$). Each participant was given the 65 tasks in a randomly generated order and was asked to complete them within an hour. Although some participants did not complete all tasks, all of them were completed by roughly the same number of participants, and the distribution was the following: 4-objects tasks ($M = 89.59$, $SD = 1.94$); 5-object tasks ($M = 85.60$, $SD = 3.07$); 6-object tasks ($M = 79.00$, $SD = 1.41$); 7-object tasks ($M = 79.10$, $SD = 3.58$); 8-object tasks ($M = 78.20$, $SD = 1.54$).

On average, participants took about 40 minutes to complete the tasks. For each task, we recorded the ordering and object placement locations inside the container. The collected dataset is grouped per task class (a task class is a set of tasks with the same number of objects). For each task class: (a) we cluster the recorded solutions with respect to their spatial patterns using Principal Components Analysis (PCA) (see); (b) we classify the provided solutions into a set of classes by looking at the first two item placements and comparing the frequency of each ordered pair. \figref{fig:results} depicts an example packing task completed by a participant, and illustrates the associated packing trends extracted with the outlined process.. \figref{fig:study_box_cluster} illustrates four distinct spatial clusters of object placements that emerged in subjects' placements. For the same task, \figref{fig:study_frequency} describes the frequency of different temporal patterns that emerged.

To extract a more holistic view of the data, we look at the distributions of placement and ordering strategies per task class. Specifically, for each task class, we count the frequency that each placement cluster occurred, and the frequency that each ordering occurred. To quantify the relative preferences of participants over different strategies, we compute the information entropy for each the frequency distributions. Intuitively, entropy quantifies how uniform the distribution was (higher entropy indicates higher uniformity in the distribution). To make sure the entropy calculations are relatable across task classes, we normalize the raw values with the maximum theoretical entropy per task class (the maximum entropy corresponds to the event that all clusters are equally represented in a task class).

\begin{figure}
\centering
\includegraphics[width=\linewidth]{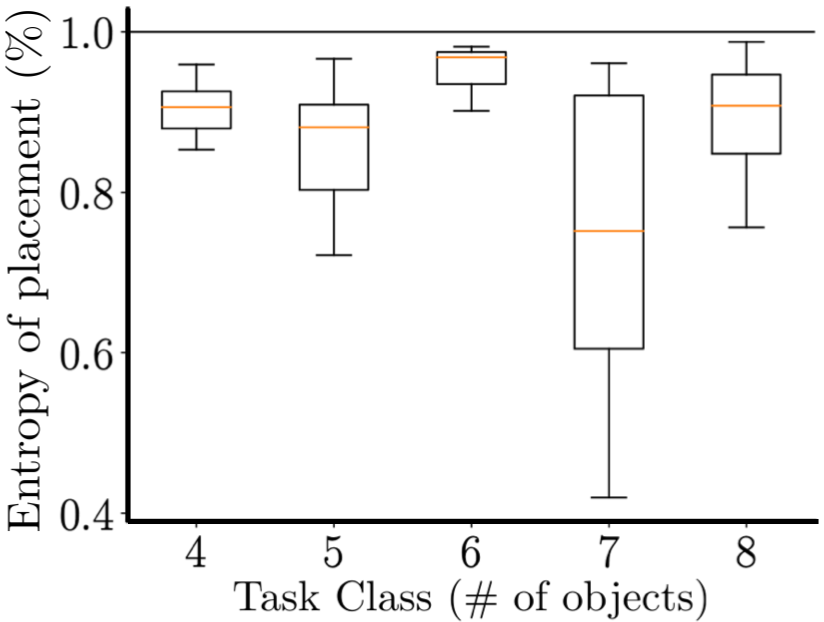}
\caption{Normalized Information Entropy of placement frequency per task. The Yellow lines indicate mean entropy per task class, whereas error bars correspond to standard deviations. The entropy stays high across the tasks, indicating that the participants did not have strong preference among the feasible spacial clusters.\label{fig:placement}}
\end{figure}

\begin{figure}
\centering
\includegraphics[width=\linewidth]{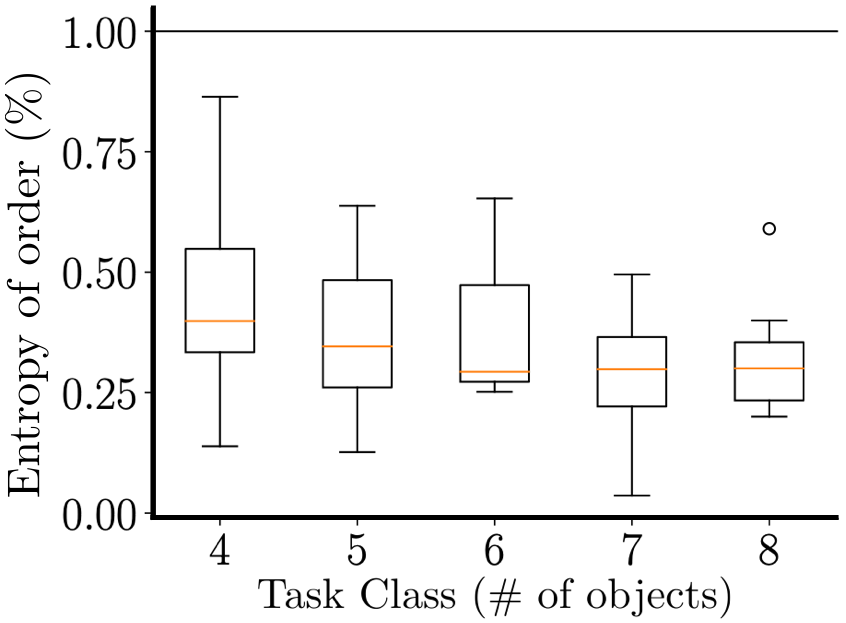}
\caption{Normalized Information Entropy of order frequency per task. The Yellow lines indicate mean entropy per task class, whereas error bars correspond to standard deviations. Even as the number of objects per task increases, the entropy stays low, indicating strong temporal patterns.\label{fig:ordering}}
\end{figure}

\figref{fig:placement} depicts the normalized entropy of the placement frequency distribution, whereas \figref{fig:ordering} shows the normalized entropy of the ordering frequency distribution. We see relatively high entropy in the placement strategies across all task classes. High entropy in this case indicates that the placement clusters tended to be roughly equally represented in participant's packing strategies. Although placement trends seem to exist, they are not as strong as we expected. Further visual inspection of the examples showed that participants tended to place larger items at the corners of the container. This could be the result of human preferences and decision making, but it could also be an artifact of the constrained design. Regarding the ordering, we see relatively low-entropy frequencies across all task classes. This indicates the existence of a temporal structure in subjects' strategies, i.e., participants were partial towards specific orderings.
Qualitatively, we have observed that participants tend to choose larger objects first, put them closer to one of the four corners, and then pack smaller items in descending order in size.

\section{Discussion}

The findings of this study illustrate our extracted knowledge about the particular domain in consideration, i.e., that of 2-dimensional packing. We discovered that human packing strategies in this domain tend to follow specific spatiotemporal patterns. These patterns are indicative of spatial and temporal preferences of humans in packing tasks. It appeared that the temporal preferences were quite strong (see \figref{fig:ordering}), whereas the spatial preferences not as much (see \figref{fig:placement}). Additional qualitative examination of the discovered patterns provided insights into the types of these preferences. Notable examples include participants' inclination towards placing larger objects in the beginning of the task, and placing larger objects at the corners of the container. We expect that identifying and adapting to observed packing strategies online could enable an artificial agent to assist a human agent effectively.

\section{Ongoing \& Planned Work}

Our key insight is that understanding the mechanisms underlying human decision making could enable an artificial agent to provide effective assistance, yielding improved task performance and reducing cognitive load for human users. Some domains can be particularly challenging for humans, for reasons related to the limits of human computational abilities. For example, in the packing domain, the limited human planning horizon and human spatial efficiency can greatly affect task performance and mentally load humans to an undesired extent. In fact, packing can be cast as the knapsack optimization problem, which is known to be NP-hard \citep{garey2002computers}. We expect that an AI agent, capable of modeling both the knapsack problem and the mechanisms underlying human decision making could provide effective assistance resulting in improved task performance and reduced cognitive load for human users. Ongoing work involves the development of a planning framework that would allow us to test this hypothesis through a follow-up user study.

\subsection{A Framework for Packing Assistance}

Motivated by the findings of the presented study, we develop a framework for planning under uncertainty that incorporates modeling of human decision making in collaborative packing tasks. In particular, we are working on adapting the Bayesian Reinforcement Learning (BRL) framework of \citet{lee2018bayesian} to enable reasoning about uncertainty over human packing strategies. BRL is a reinforcement learning framework that incorporates a mechanism for reasoning about model uncertainty. It models the problem as a Bayes-Adaptive Markov Decision Process (BAMDP) \citep{Duff02}, explicitly modeling uncertainty as a belief over a latent uncertainty variable, incorporated in the transition function and the reward function. Overall, BRL maximizes the expected discounted reward, given the uncertainty. We believe that this mechanism is of particular relevance and value in problems involving human interaction, where uncertainty is typically over human mental models underlying their decision making.

For our task domain, we are incorporating a belief distribution over the human user's spatiotemporal placement strategy, given the container configuration and the object's shape. We plan on using the collected human dataset to learn the outlined predictive model. During execution, we will be using our framework as a recommender system that will be providing online recommendations to the human user.

\subsection{Planned User Study}

To formally investigate our outlined insight, we design an online user study, in which human subjects will be exposed to a set of conditions (within-subjects), corresponding to different modes of AI assistance. More specifically, we consider the following set of conditions:
\begin{enumerate}
\item No recommendation -- the user completes the task without receiving any assistance.
\item The system provides object recommendations, i.e., assists by manipulating the order of object placements.
\item The system provides both order and placement recommendations.
\item The system provides random object recommendations.
\item The system provides random order and random placement recommendations.
\end{enumerate}

We hypothesize that the assistive conditions will yield improved task performance compared to the condition of no assistance, but also more positive human ratings and reduced reported cognitive load. As performance metrics, we consider the time-to-completion and the packing spatial efficiency. After each condition, we will collect ratings of perceived system intelligence, likeability, and predictability, based on the Godspeed \citep{bartneckHRIMetrics2009} to understand the perception of the considered conditions from the perspective of participants. Finally, we will measure the cognitive load associated with each condition by presenting a questionnaire based on the NASA-TLX \citep{nasatlx}. Finally, participants will be provided with an open-form question, asking them to provide qualitative feedback of their choice regarding their interaction with the system.

\section{Acknowledgments}
Gilwoo Lee is partially supported by Kwanjeong Educational Foundation. This work was partially funded by the Honda Research Institute USA, the National Science Foundation NRI (award IIS-1748582), and the Robotics Collaborative Technology Alliance (RCTA) of the United States Army Laboratory.

\balance
\bibliography{references}
\bibliographystyle{aaai}

\end{document}